\pdfoutput=1

\documentclass[11pt]{article}

\usepackage{EMNLP2023}

\usepackage{times}
\usepackage{latexsym}

\usepackage[T1]{fontenc}

\usepackage[utf8]{inputenc}

\usepackage{microtype}

\usepackage{inconsolata}
\usepackage{amssymb}
\usepackage{amsmath}
\usepackage{url}
\usepackage{subcaption}

\usepackage{textcomp}
\usepackage{graphicx}
\usepackage{tikz}
\usepackage{arydshln}
\usepackage{array} 
\usepackage{multirow}
\usepackage{listings}
\usepackage{xcolor}
\usepackage{longtable}
\usepackage{examples}

\usetikzlibrary{calc}

\usepackage{filecontents}
\DeclareMathOperator*{\concat}{\scalebox{1}[1.5]{$\parallel$}}

\usepackage{floatrow}
\newfloatcommand{capbtabbox}{table}[][\FBwidth]

\usepackage{blindtext}

\usepackage{tabularx}
\usepackage{booktabs}
\usepackage{colortbl}

\usepackage{pgfplots}
\pgfplotsset{compat=1.12}

\title{Conversational Semantic Parsing using Dynamic Context Graphs}

\author{Parag Jain \qquad Mirella Lapata \\
Institute for Language, Cognition and Computation\\
School of Informatics, University of Edinburgh\\
 10 Crichton Street, Edinburgh EH8 9AB\\
\texttt{parag.jain@ed.ac.uk}~~~~\texttt{mlap@inf.ed.ac.uk}\\
}
\begin{document}
\maketitle

\begin{abstract}
In this paper we consider the task of conversational semantic parsing
over general purpose knowledge graphs (KGs) with millions of entities,
and thousands of relation-types. We focus on models which are capable
of interactively mapping user utterances into executable logical forms
(e.g., \textsc{Sparql}) in the context of the conversational history.
Our key idea is to represent information about an utterance and its
context via a subgraph which is created dynamically, i.e.,~the number
of nodes varies per utterance.  Rather than treating the subgraph as a
sequence, we exploit its underlying structure and encode it with a
graph neural network which further allows us to represent a large
number of (unseen) nodes. Experimental results show that dynamic
context modeling is superior to static approaches, delivering
performance improvements across the board (i.e., for simple and
complex questions). Our results further confirm that modeling the
structure of context is better at processing discourse information,
(i.e., at handling ellipsis and resolving coreference) and longer
interactions.

\end{abstract}
\section{Introduction}

General purpose knowledge graphs (KG), like
Wikidata~\citep{10.1145/2629489} structure information in a semantic
network of entities, attributes, and relationships, allowing machines
to tap into a vast knowledge base of facts. Knowledge base question
answering (KBQA) is the task of retrieving answers from a KG, given
natural language questions. A popular approach to KBQA (see
\citealt{gu2022knowledge} and the references therein) is
based on semantic parsing which maps questions to logical form queries
(e.g., in \textsc{Sparql}) that return an answer once executed against the KG.

\begin{figure}[t]
  \centering
  \small
\begin{tabular}{@{}l@{~}p{7cm}} \toprule
1. & \textcolor{blue}{Who starred in Mathias Kneissl ?} \\
& \tt{SELECT ?x WHERE \{ wd:Q3298576 wdt:P161 ?x . ?x wdt:P31 wd:Q502895 .  \}} \\
 & \textcolor{cyan}{Rainer Werner Fassbinder, Volker Schlöndorff, Hanna Schygulla} \\
\hline
2. & \textcolor{blue}{Who was the director of that work of art ?} \\
 &\tt{SELECT ?x WHERE \{ wd:Q3298576 wdt:P57 ?x . ?x wdt:P31 wd:Q502895 .  \}} \\
 & \textcolor{cyan}{Reinhard Hauff} \\
\hline
3. & \textcolor{blue}{Does Dubashi have that person as actor ?} \\
& \tt{ASK \{ wd:Q76025 wdt:P161 wd:Q24807818 .  \}} \\
 &\textcolor{cyan}{No} \\
\hline
4.  &\textcolor{blue}{Which works of art are Rainer Werner Fassbinder or Laura Esquivel a screenwriter of ?} \\
 & \tt{SELECT ?x WHERE \{ \{ ?x wdt:P58 wd:Q44426 . ?x wdt:P31
   wd:Q838948 .  \}} \\
 & \tt{UNION \{ ?x wdt:P58 wd:Q230586 . ?x wdt:P31 wd:Q838948 .  \} \}}\\
 & \textcolor{cyan}{The American Soldier, Lili Marleen, Love Is Colder Than
Death ...}\\ 
\hline
&\textcolor{darkgray}{\tt{Q3298576: Mathias Kneissl, Q76025: Reinhard Hauff, Q24807818: Dubashi, Q44426: Rainer Werner Fassbinder
Q230586: Laura Esquivel, Q838948: work of art, Q502895: common name, P161: cast member, P31: instance of, P57: director, P58: screenwriter}}\\
\bottomrule
\end{tabular}
\caption{Example interaction from $\mathbb{SPICE}$ dataset
  \cite{spice} with \textcolor{blue}{utterances}, corresponding
  \textsc{Sparql} queries, and \textcolor{cyan}{answers} returned
  after executing the queries on the Wikidata graph engine. The bottom
  block shows the KG elements (i.e.,~graph nodes) involved in this
  interaction.}
\label{fig:example_interaction}
\end{figure}

Existing work
(e.g.,~\citealt{bogin-etal-2019-global,ravishankar2021twostage,YIN2021510})
has mostly focused on answering questions in isolation, whereas we
consider the less studied task of \emph{conversational} semantic
parsing.  Specifically, our interest lies in building systems capable
of interactively mapping user utterances to executable logical forms
in the \emph{context} of previous
utterances. Figure~\ref{fig:example_interaction} shows an example of a
user-system interaction, taken from $\mathbb{SPICE}$~\citep{spice}, a
recently released conversational semantic parsing dataset. Each
interaction consists of a series of utterances that form a coherent
discourse and are translated to executable semantic parses (in this
case \textsc{Sparql} queries).  Interpreting each utterance, and
mapping it to the correct parse needs to be situated in a particular
context as the exchange proceeds.  To answer the question in utterance
2, the system needs to recall that \textsl{Mathias Kneissl} is still
the subject of the conversation, however, the user is no longer
interested in who starred in the film but in who directed it. It is
also natural for users to omit previously mentioned information (e.g.,
through ellipsis or coreference), which would have to be resolved to
obtain a complete semantic parse.

In addition to challenges arising from processing contextual
information, the semantic parsing task itself involves linking
entities, types, and predicates to elements in the KG
(e.g.,~\textsl{Mathias Kneissl} to~{\tt Q3298576}) whose topology is
often complex with a large number of nodes. Moreover, unlike
relational databases, the schema of an entity is not static but
dynamically instantiated~\citep{gu2022knowledge}. For example, the
entity type \verb|person| can have hundreds of relations but only a
fraction of these will be relevant for a specific
utterance. Therefore, to generate faithful queries, we cannot rely on
memorization and should instead make use of local schema
instantiation. In general, narrowing down the set of entities and
relations is critical to parsing utterances requiring complex
reasoning (i.e.,~where numerical and logical operators apply over sets
of entities).

Existing work~\citep{spice} handles the aforementioned challenges by
adopting various simplifications and shortcuts. For instance, since it
is not feasible to encode the entire KG, only a subgraph relevant to
the current utterance is extracted and subsequently linearized and
treated as a sequence.  Entity type information that is not directly
accessible via neighboring subgraphs is obtained through a
\textit{global} lookup (essentially a reverse index of all types in
the KG). This solution is computationally expensive, as the lookup is
performed practically for every user utterance, and does not scale
well (the index would have to be recreated every time the KG changed).

In this paper we propose a modeling approach to conversational
semantic parsing which relies on \emph{dynamic context graphs}. Our
key idea is to represent information about an utterance and its
context through a dynamically generated subgraph, wherein the number
of nodes varies for each utterance. Moreover, rather than treating the
subgraph as a sequence, we exploit its underlying structure and encode
it with a graph neural network \cite{4700287,1555942}.  To improve
generalization, we learn \textit{implicit} node embeddings by
aggregating information from neighboring nodes whose embeddings are in
turned initialized through a pretrained
model~\citep{devlin-etal-2019-bert}. In addition, we introduce
context-dependent type linking, based on the entity and its
surrounding context which further helps with type disambiguation.

Experimental evaluation on the $\mathbb{SPICE}$ dataset \cite{spice}
demonstrates that modeling context dynamically is superior to static
approaches, improving performance across the board (i.e., for simple
and complex questions requiring comparative or quantitative
reasoning). Our results further confirm that modeling the structure of
context is better at processing discourse information, (i.e., at
handling ellipsis and resolving coreference) and longer interactions
with multiple turns.

\section{Related Work}
Previous work on semantic parsing for KBQA~\cite{gu2022knowledge} has
focused on mapping stand-alone utterances to logical form
queries. Various approaches have been proposed to this effect which
broadly follow three modeling paradigms. Ranking methods first
enumerate candidate queries from the KB and then select the query most
similar to the utterance as the semantic parse
\cite{ravishankar2021twostage, 10.1007/978-3-030-88361-4_8,
  10.1145/3357384.3358033}. Coarse-to-fine methods
\citep{dong-lapata-2018-coarse,ding-etal-2019-leveraging,ravishankar2021twostage}
perform semantic parsing in two stages, by first predicting a query
sketch, and then filling in missing details. Finally, generation
methods \citep{YIN2021510} first rank candidate parses and then
predict the final parse by conditioning on the utterance and best
retrieved logical forms.

Our task is most related to conversational text-to-SQL parsing, as
manifested in datasets like SParC~\citep{yu-etal-2019-sparc},
CoSQL~\citep{yu-etal-2019-cosql}, and
ATIS~\citep{dahl-etal-1994-expanding, suhr-etal-2018-learning}.  SParC
and CoSQL cover multiple domains and include multi-turn user and
system interactions. These datasets are challenging in requiring
generalization to unseen databases, but the conversation length is
fairly short and the databases relatively small-scale. ATIS contains
utterances paired with SQL queries pertaining to a US flight booking
task; it exemplifies several long-range discourse
phenomena~\citep{10.1162/tacl_a_00422}, however, it covers a
single domain with a simple database schema.

Graph-based methods have been previously employed in semantic parsing
primarily to  encode the database
schema, so as to  enable the parser to globally reason about
the structure of the output query \cite{bogin-etal-2019-global}.

Other work \citep{hui-etal-2022-s2sql} uses relational graph networks
to jointly represent the database schema and syntactic dependency
information from the questions. In the context of conversational
semantic parsing, \citet{cai-wan-2020-igsql} use a graph encoder to
model how the elements of the database schema interact with
information in preceding context. Along similar lines,
\citet{Hui:ea:2021}, use a graph neural network with a dynamic memory
decay mechanism to model the interaction of the database schema and
the utterances in context as the conversation proceeds.  All these
approaches encode the schema of relational databases which are
significantly smaller in size (e.g., number of entities and types of
relations) compared to large-scale KGs, where encoding the entire
graph in memory is not feasible.

Closest to our work are methods which cast conversational KGQA as a
semantic parsing task~\citep{kacupaj-etal-2021-conversational,
  marion-etal-2021-structured}. These approaches build hand-crafted
grammars that are not directly executable to a KG engine. Furthermore,
they assume the KG can be fully encoded in memory which may not be
feasible in real-world settings. \citet{spice} develop a parser which
is executable with a real KG engine (e.g., Blazegraph) but simplify
the task by considering only limited conversation context.

\section{Problem Formulation}
Given a general purpose knowledge graph, such as Wikidata, our task is
to map user utterances to formal executable queries, \textsc{Sparql}
in our case. We further assume an \emph{interactive} setting where
users converse with the system in natural language and the system
responds while taking into account what has already been said (see
Figure~\ref{fig:example_interaction}).  The system's response is obtained
upon \emph{executing} the query against a graph query engine.

Let $G$ denote the underlying KG and $I$~a single
interaction. $I$~consists of a sequence of turns where each turn is
represented by $\langle X_t, A_t, Y_t \rangle$ denoting an
utterance-answer-query triple at time~$t$ (see blocks in
Figure~\ref{fig:example_interaction}). A user utterance~$X_t$ is a
sequence of tokens $\langle x_1, x_2, \dots ,x_{|X_t|} \rangle$,
where~$|X_t|$ is the length of the sequence and each $x_i, i \in [1,
  |X_t|]$ is a natural language token. Query string $Y_t$ is a
sequence of tokens $\langle y_1, y_2, \dots ,y_{|Y_t|} \rangle$,
where~$|Y_t|$ is the length of the sequence and each $y_i, i \in [1,
  |Y_t|]$ is a either a token from the \textsc{Sparql} syntax vocabulary
(e.g., {\tt SELECT, WHERE}) or a KG element $\in G$ (e.g.,~{\tt
  Q3298576}). Answer~$A_t$ at time~$t$ is the result of
executing~$Y_t$ against~$G$. Given the interaction history~$I[:t - 1]$
at turn~$t$ and current utterance~$X_t$, our goal is to
generate~$Y_t$.  This involves understanding~$X_t$ in the context
of~$X_{:t - 1}$, $A_{:t-1}$, and~$G$, and learning to generate $Y_t$
based on encoded contextual information.

\section{Model}
\label{sec:model}

\begin{figure*}[t!]
\begin{center}
\includegraphics[scale=0.82]{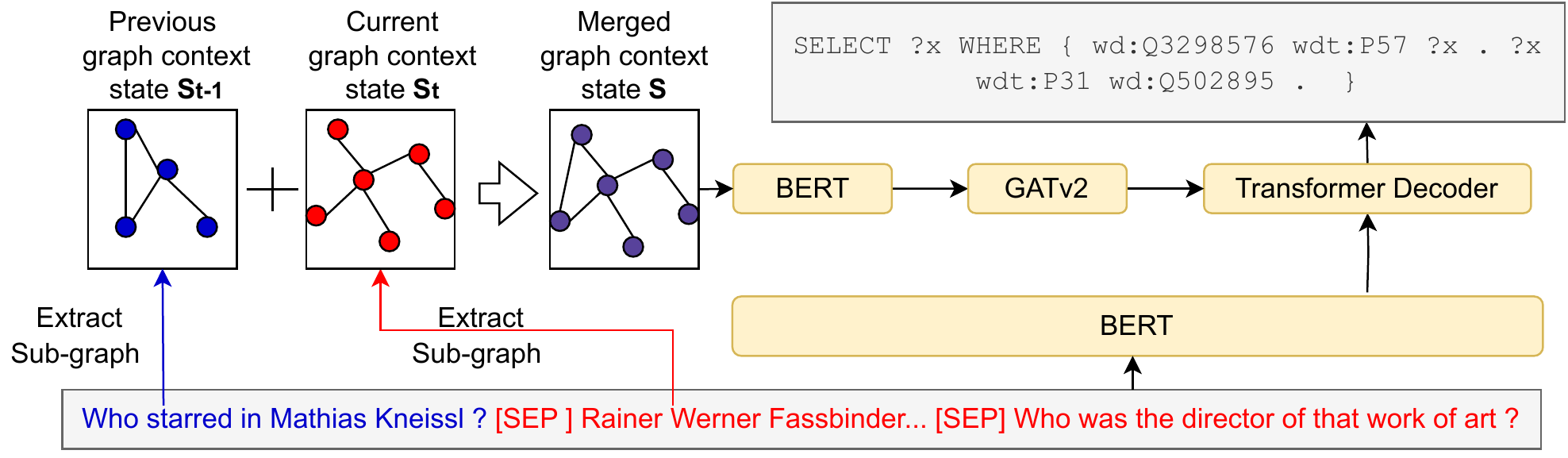}
\caption{Model architecture. The \textcolor{blue}{previous} and
  \textcolor{red}{current} utterance are concatenated and their
  subgraphs are \textcolor{violet}{merged} and encoded in a graph
  neural network. The subgraphs represent the entity neighborhood and
  type linking.}
\label{fig:model}
\end{center}
\end{figure*}

Our modeling approach combines three components. We first ground
named entities in the user utterance to KG entities, and use these
linked entities to extract a subgraph that functions as context
(Section~\ref{sec:entity:grounding}). The second component is
responsible for type linking in the context of current and previously
mentioned named entities (Section~\ref{sec:type:linking}). And
finally, our semantic parser learns to map user utterances into
\textsc{Sparql} queries (Section~\ref{sec:dcg}). 

\subsection{Entity Grounding and Disambiguation}
\label{sec:entity:grounding}

We are interested in grounding user utterance~$X_t$ to
graph~$G$. Since encoding the entire KG is not feasible, we extract a
subgraph from~$G$ which is relevant to the current turn. To achieve
this, we first perform named entity recognition with an off-the-shelf
NER system, in our experiments we use
AllenNLP~\citep{Gardner2018AllenNLP}. We perform named entity linking
through efficient string matching\footnote{Our implementation is based
  on pyahocorasick \url{https://pypi.org/project/pyahocorasick/}.}
\cite{10.1145/360825.360855} unlike \citet{spice} who deploy an
ElasticSearch server for querying an inverted index. 

A string can be ambiguous, i.e.,~link to multiple entities. For
example, \textsl{Rainer Werner Fassbinder} can be linked to
\textsl{filmmaker} ({\tt Q44426}) and \textsl{movie} ({\tt
  Q33561976}). To deal with ambiguity and to increase recall,
\citet{spice} do not ever commit to a single entity but instead
include the top-$K$ matching ones; however, this introduces noise and
increased computational cost. Instead, we disambiguate entities based
on their popularity in the training set~\citep{6823700} and compare
the two approaches in Section~\ref{sec:results}.

Following~\citet{spice}, for each identified KG entity~$e$, we extract
triples~$(e,r,o)$ and~$(s,r,e)$, where $s$ and $o$ denote the subject
or object of relation~$r$. For instance, entity \textsl{Dubashi} would
have triple \textsl{(Dubashi, country of origin, India)}. Subjects and
objects are further mapped to their types in place of actual entities
(i.e.,~$(e , r, o_{type})$ and $(s_{type},r,e)$). In our example, the
triple for \textsl{Dubashi} then becomes \textsl{(Dubashi, country of
  origin, country)}, where \textsl{country} is type {\tt Q6256}. We
denote the set of typed triples as~$G^{ent}_{t}$.

\subsection{Context-Dependent Type Linking}
\label{sec:type:linking}

Entities in \textsc{Sparql} queries have types, for example, in
Figure~\ref{fig:example_interaction}, KG element {\tt Q502895} is a
placeholder for the type ``{common name}''. Type instances are often
present in the one-hop entity neighborhood~$G^{ent}_{t}$, but can also
be more hops away.  \citet{spice} index \emph{all} KG types and
perform a \textit{global} lookup which is computationally expensive,
and solely applicable to the KG they are working with. Instead, we
perform type linking based on the entities mentioned in the
\textit{current} context. We expand the grounded entities to extract
triples with type information.\footnote{For entity \textit{ent} we
  query \tt{select ?r1 ?n1 ?t1 ?r2 ?n2 ?t2 where \{ wd:ent ?r1 ?n1
    . ?n1 wdt:P31 ?t1 . OPTIONAL \{?n1 ?r2 ?n2 . ?n2 wdt:P31 ?t2\}
    \}}.} Since considering multi-hop neighborhoods would lead to
extremely large subgraphs and would not be memory efficient, we prune
these based on their string overlap with the user utterance,
significantly reducing the number of triples. The pruned
graph~$G^{type}_{t}$ is merged with the previously obtained entity
graph~$G^{ent}_{t}$, such that, $G_{t} = G^{ent}_{t} \cup
G^{type}_{t}$.

\subsection{Dynamic Context Graph Model}
\label{sec:dcg}

Figure~\ref{fig:model} shows the overall architecture of our dynamic
context graph model (which we abbreviate to DCG). DCG takes as input a
tuple of form $\langle C_t, X_t, G_{t} \rangle$, where~$X_t$ is a user
utterance at time~$t$ and $G_t$ is the corresponding
subgraph. $C_t$~denotes the previous context information that includes
the previous utterance $X_{t -1}$, the previous answer $A_{t - 1}$,
and subgraph $G_{t - 1}$. We use~$\hat{G_t}$ to represent merged
context subgraphs $G_t$ and $G_{t - 1}$ such that $\hat{G_t} = G_{t}
\cup G_{t - 1}$. We encode the context subgraph~$\hat{G_t}$ with a
graph neural network (GNN, \citealt{4700287,1555942}) and user
utterances and their discourse context with
BERT~\citep{devlin-etal-2019-bert}. Our decoder is a transformer
network~\citep{NIPS2017_3f5ee243} that conditions on the user
utterance encoding and the corresponding graph representations.

\paragraph{Utterance Encoder} We use BERT\footnote{We employ BERT for
 a fair comparison with prior work. Nonetheless, our model does not
 have any inherent restrictions that would prevent the use of other
 pretrained models.} \citep{devlin-etal-2019-bert} to represent the
concatenation of previous utterance~$X_{t - 1}$, previous answer~$A_{t
  - 1}$, and current utterance~$X_t$ (see Figure~\ref{fig:model}). To
distinguish between current and past context we use the \textsc{[SEP]}
token. More formally, let
$\hat{X_t}~=~$\textsc{[CLS]}~$X_{t - 1} $~\textsc{[SEP]}$~A_{t -
  1}~$\textsc{[SEP]}~$X_t$ denote the input to BERT,
where \mbox{$X_t = (x_1, x_2 \dots x_{|X_t|})$}, $A_{t - 1} = (a_1, a_2 \dots a_{|A_{t-1}|})$, and $X_{t - 1} = (x_1,
x_2 \dots x_{|X_{t - 1}|})$ are sequences of natural language
tokens. We obtain latent representations $Z_t$ as $Z_t =
\operatorname{BERT}(\hat{X_t})$.

\paragraph{Graph Encoder} We represent the KG subgraph~$\hat{G_t}$ at
time~$t$ as a directed graph~\mbox{$\mathcal{G} = (\mathcal{V},
  \mathcal{E})$} (hereafter, we simplify notation and drop time~$t$),
where $\mathcal{V} = \{v_1, v_2, \dots , v_n\}$ are nodes such that
\mbox{$v_i \in \{entities,~relations,~types\}$} and $\mathcal{E}
\subseteq \mathcal{V} \times \mathcal{V} $. Each node $v_i$ consists
of a sequence of natural language tokens, such that \mbox{$v_i = \langle
v_{i1}, v_{i2}, \dots ,v_{i|v_i|} \rangle$}. Our KG has a large number
of distinct nodes, but we cannot possibly attest all of them during
training. To handle unseen nodes at test time, we obtain a
generic node representation~$h_{i}^{0}$ for node~$v_i$, where
$h_{i}^{0} = \operatorname{AVG}(\operatorname{BERT}(v_i))$. In other
words, we compute encoding~$h_{i}^{0}$ by taking the average of the
individual token encodings obtained from BERT. We do not create a
separate embedding matrix but directly update the BERT representations
during learning, which allows us to scale to a large number of
(unseen) nodes.

A graph neural network (GNN) learns node representations by
aggregating information from its neighboring nodes. Each GNN
layer~$l$, takes as input node representations $\{h_{i}^{l - 1} \mid i
\in [1, n]\}$ and edges~$\mathcal{E}$. The output of each layer is an
updated set of node representations $\{h_{i}^{l} \mid i \in [1,
  n]\}$. We use Graph Attention Network v2 (GATv2,
\citealt{brody2022how}) for updating node representations which
replaces the static attention mechanism of GAT~\cite{gat2018graph}
with a dynamic and more expressive variant.  Let $\mathcal{N}_i = \{
v_j \in \mathcal{V} \mid \bigl( j,i \bigr) \in \mathcal{E} \} $ denote
the neighbors of node $v_i$ and~$\alpha_{ij}$ the attention score
between node~$h_i$ and~$h_j$. We calculate attention as a single-layer
feedforward neural network, parametrized by a weight vector $a$ and
weight matrix $W$:
\begin{equation}\label{attncoeff}
    \alpha_{ij} = \frac{\exp\bigl( \psi \bigl( h_i^{l - 1}, h_j^{l - 1} \bigr) \bigr)}{\sum_{k\in\mathcal{N}_i}\exp\bigl(\psi\bigl(h_i^{l - 1}, h_k^{l - 1}\bigr)\bigr)}
\end{equation}
The scoring function $\psi$ is computed as follows:
\begin{multline}
     \psi \bigl( h_i^{l - 1} , h_j^{l - 1} \bigr) = \\ a^T\operatorname{LeakyReLU}\bigl( W \cdot [h_i^{l - 1} \parallel h_j^{l - 1}]\bigr)
     \end{multline}
where~$\cdot^T$ represents transposition and~$\parallel$ is the
concatenation operation. Attention coefficients corresponding to each
node~$i$ are then used to compute a linear combination of the features
corresponding to neighboring nodes as:
\begin{equation}\label{eqnatt}
	h_i^{l} = \sigma\left(\sum_{j\in\mathcal{N}_i} \alpha_{ij} {W}h_j^{l - 1}\right)
\end{equation}

\paragraph{Decoder} Our decoder is a transformer
network~\citep{NIPS2017_3f5ee243}. Let $ H^{l}_t = \bigl( h_{1t}^{l},
h_{2t}^{l}, \dots , h_{nt}^{l} \bigr)$ denote the sequence of node
representations from the last layer of the graph network (recall
$t$~here represents an interaction turn). Our decoder models the
probability of generating a \textsc{Sparql} parse conditioned on the
graph and input context representations, i.e., $p(Y_t \mid H^l_t,
Z_t)$.  Generating the \textsc{Sparql} parse requires generating
syntax symbols (such as {\tt SELECT, WHERE)} and KG elements (i.e.,
entities, types, and relations). Given  decoder state~$s_i$ at the
$i^{th}$ generation step, the probability of generating~$y_i$ is
calculated as:
\begin{multline}
    p(y_i \mid y_{<i}, H^l_t, s_i) = p_{\mathcal{G}}(y_i \mid y_{<i}, H^l_t, s_i) 
    \\ + p_{\mathcal{S}}(y_i \mid y_{<i}, s_i)
\end{multline}
where $p_{\mathcal{G}}$ and $p_{\mathcal{S}}$ are the probability of
generating a graph node and syntax symbol, respectively. We calculate
$p_{\mathcal{S}} = \operatorname{softmax}(W_{1} s_{i})$, such that
\mbox{$W_{1} \in \mathbb{R}^{|V_{s}| \times d} $}, and~$|V_{s}|$ is
the \textsc{Sparql} syntax vocabulary size. We
calculate~$p_{\mathcal{G}}$ using node embeddings~$H^l_t$, as~$p_{\mathcal{G}} = \operatorname{softmax}(H^l_t s_{i})$.

\paragraph {Training}
Our model is trained end-to-end by optimizing the cross-entropy
loss. Given  training instance $\langle C_t, X_t, Y_t , G_{t}
\rangle$, where $Y_t$ is a sequence of gold output tokens $\langle
y_1, y_2, \dots ,y_{|Y_t|} \rangle$, we minimize the token-level
cross-entropy as:
\begin{equation}
    \mathcal{L}(\hat{y}_i) = - logp(y_i \mid X_t, G_t, C_t)
\end{equation}
where~$\hat{y}_i$ denotes the predicted output token at decoder step~$i$. 

\section{Experimental Setup}

\paragraph{Dataset}
We performed experiments on $\mathbb{SPICE}$ \cite{spice}, a recently
released large-scale
dataset\footnote{\url{https://github.com/EdinburghNLP/SPICE}.} for
conversational semantic parsing built on top of the CSQA benchmark
\cite{csqa}.  $\mathbb{SPICE}$ consists of user-system interactions
where natural language questions are paired with \textsc{Sparql}
parses and answers provided by the system correspond to
\textsc{Sparql} execution results (see
Figure~\ref{fig:example_interaction}). We present summary statistics
of the dataset in Table~\ref{tab:statistics:spice}. As can be seen, it
contains a large number of training instances, the conversations are
relatively long (the average turn length is~9.5), and the underlying KG
is sizeable (12.8M entities). $\mathbb{SPICE}$ has simple factual
questions but also more complicated ones requiring reasoning over sets
of triples; it also exemplifies various discourse-related phenomena
such as coreference and ellipsis. We provide examples of the types of
questions attested in $\mathbb{SPICE}$ in
Appendix~\ref{sec:app:ques_types}.

\begin{table}[t]
\centering
{\small
  \begin{tabular}{lc}  \toprule
  
  \# instances & 197K \\
  \# entities & 12.8M\\
  \# relations & 2,738\\
  \# types & 3,064\\
  Avg. turn length & 9.5\\
  Avg. entities per conversation & 7.6\\
  Avg. types per conversations & 6.5\\
  Avg. neighborhood per turn & 181.4 triples\\ \bottomrule
  \end{tabular}
}
\caption{Statistics of the $\mathbb{SPICE}$ dataset.} \label{tab:statistics:spice}
\end{table}


\paragraph{Evaluation Metrics}
Following previous work~\citep{spice}, we report exact match accuracy
and F1 (or accuracy depending on question type). Exact match is the
percentage of predicted \textsc{Sparql} queries that string match with
the corresponding gold \textsc{Sparql}. F1 (or answer accuracy) is
calculated between execution results of predicted queries and gold
queries. For Verification queries and queries involving Quantitative
and Comparative Reasoning, we calculate execution answer accuracy.
For other types of questions, F1 scores are calculated by treating the
results as a set.

\paragraph{Model Configuration}
\label{sec:model_config}
Our model is implemented using PyTorch~\citep{NEURIPS2019_bdbca288}
and trained with the AdamW~\citep{loshchilov2018decoupled}
optimizer.\footnote{Our code can be downloaded from
  \url{https://github.com/parajain/dynamic_context}.} Model selection was based on exact match accuracy on the
validation set. We used two decoder layers and two GATv2 layers for
all experiments. We used HuggingFace's pretrained BERT embeddings
\citep{wolf-etal-2020-transformers}, specifically the uncased base
version.  Our GATv2 implementation is based on PyTorch
Geometric~\citep{Fey/Lenssen/2019} with two attentions heads. We use
adjacency matrices stacking as a method of creating mini-batches for
our GNN across different examples. We identify named entities using
the AllenNLP named entity recognition (NER) system
\cite{Gardner2018AllenNLP}. Our execution results are based on the
Wikidata subgraph provided by~\citet{spice}. Our SPARQL server is
deployed using Blazegraph.\footnote{\url{https://blazegraph.com/}} See
Appendix~\ref{sec:app:model_details} for more implementation details.

As described in Section~\ref{sec:dcg}, at each utterance, our model
encodes the previous~$t$ subgraphs. Larger context is informative but
can also introduce noise. We treat~$t$ as a hyperparameter and
optimize its value on the development set. We report results
with~$t=5$ (see Appendix~\ref{appendix:context_len_appendix} for
details).

\paragraph{Comparison Models}
\label{systems}

We compare against the semantic parser of \citet{spice}. Their model
is based on BERT \cite{devlin-etal-2019-bert}, it relies on AllenNLP
\cite{Gardner2018AllenNLP} for named entity recognition, and performs
entity linking with an
ElasticSearch\footnote{\url{https://www.elastic.co/}} inverted
index. As mentioned earlier, they do not explicitly perform named
entity disambiguation (they consider the $K=5$ best matching entities
and their neighborhood graphs as part of the vocabulary) and use a
{global lookup} for type linking. As the size of the linearized
subgraph often exceeds BERT's maximum number of input tokens (which is
512), they adopt a workaround where the graph is chunked into several
subsequences, and encoded separately.  We refer to their Semantic
Parser as BertSP$_{\mathcal{GL}}$, where $\mathcal{GL}$ is a shorthand
for Global Lookup.

In addition to our full dynamic context graph model which performs
Context-dependent Type Linking (DCG$_{\mathcal{CL}}$), we also build a
simpler variant (DCG) which only relies on the entity neighborhood
subgraph for type information. Moreover, we create two variants of our
model, one which disambiguates entities, and another one which does
not (similar to \citealt{spice}).

\begin{figure*}[t]
\small
  \capbtabbox{%
\resizebox{\textwidth}{!}{%
 \begin{tabular}{l|cc|cc|cc|cc|cc}
    \toprule
    \multicolumn{1}{c}{~} &    \multicolumn{6}{c}{Without Disambiguation}  & \multicolumn{4}{c}{With Disambiguation}  \\
 \multicolumn{1}{c}{~} &   \multicolumn{2}{c}{DCG$_{\mathcal{CL}}$} & \multicolumn{2}{c}{DCG$_{\mathcal{GL}}$} & \multicolumn{2}{c|}{BertSP$_{\mathcal{GL}}$} & \multicolumn{2}{c}{DCG$_{\mathcal{CL}}$}  &  \multicolumn{2}{c}{DCG}  \\
        Question Type & F1 & EM & F1 & EM & F1  & EM & F1  & EM & F1  & EM \\
    \hline
    Clarification                 & 78.60 & 73.63   & 80.42   & 69.47   & \textbf{83.91} & 76.58 & 82.01   & 74.82  & 82.03  & 72.10  \\
    Logical Reasoning             & 64.12 & 49.51   & 51.14   & 31.54   & 22.74   & 28.61 & \textbf{93.95} & 79.52  & 93.33  & 78.19  \\
    Quantitative Reasoning        & 55.66 & 26.29   & \textbf{93.25} & 76.88   & 76.20   & 59.01 & 59.83   & 31.17  & 56.66  & 28.66 \\
    Comparative Reasoning         & 76.06 & 35.59   & 80.59   & 47.28   & 69.56   & 39.37 &\textbf{90.91}  & 62.46  & 90.09  & 61.11  \\
    Simple Question (Coref)       & 86.36 & 72.03   & 84.92   & 67.15   & 76.51   & 58.83 &\textbf{88.49}  & 79.90  & 87.41  & 79.18  \\
    Simple Question (Direct)      & 87.29 & 71.10   & 83.24   & 65.89   & 71.43   & 58.71 &\textbf{88.27}  & 62.25  & 85.60  & 61.44  \\
    Simple Question (Ellipsis)    & 65.22 & 56.08   & 52.84   & 47.48   & 58.14   & 50.90 & 79.08   & 83.87  & \textbf{84.35}& 82.45  \\
    \hline
     & AC & EM & AC & EM & AC & EM  & AC & EM & AC & EM \\
    \hline
    Verification (Boolean)        & 78.30 & 36.97 & 69.07 & 21.00 & 37.16 & 24.90 & \textbf{87.41} & 66.32 & 86.75 & 63.66 \\
    Quantitative Reasoning (Count)& 61.10 & 56.94 & 66.59 & 62.70 & 50.86 & 48.44 & \textbf{75.20} & 70.84 & 72.96 & 69.02 \\
    Comparative Reasoning (Count) & 42.79 & 30.40 & 60.91 & 44.68 & 43.48 & 40.67 & \textbf{67.70} & 57.34 & 66.76 & 56.60 \\
    \hline
    Overall &   69.55 & 50.85 & 72.30 & 53.41 &  59.00 & 48.60 & \textbf{81.28} & 66.85 & 80.59 & 65.24 \\
    \toprule
    \end{tabular}}
}{%
    \caption{Results on $\mathbb{SPICE}$ dataset (test
      set). BertSP$_{\mathcal{GL}}$ \cite{spice} uses NER based on
      AllenNLP) and global look-up (subscript $_{\mathcal{GL}}$) for
      type linking. DCG$_{\mathcal{CL}}$ uses context-dependent type
      linking (subscript~$_{\mathcal{CL}}$) and also AllenNLP. DCG has
      no type linking. We measure F1, Accuracy (AC), and Exact Match
      (EM). \label{tab:mainresult} } }
\end{figure*}

\section{Results}
\label{sec:results}

In this section, we evaluate the performance of our semantic parser on
the $\mathbb{SPICE}$ test set. We report results on individual
question types and overall. We also analyze our system's ability to
handle different discourse phenomena like ellipsis and coreference as
well as interactions of varying length.

\paragraph{The Effect of Dynamic Context} 
Table~\ref{tab:mainresult} summarizes our results. We first
concentrate on model variants \emph{without entity disambiguation} for
a fair comparison with \citet{spice}.

We compare DCG$_{\mathcal{GL}}$, a version of our model which adopts a
global lookup for type liking similar to BertSP$_{\mathcal{GL}}$ and
differs only in how contextual information is encoded. As we can see,
our graph-based model performs better, reaching an F1 score of~72.3\%
compared to~59\% obtained by BertSP$_{\mathcal{GL}}$ which is limited
by the way it encodes contextual information. Recall, that
BertSP$_{\mathcal{GL}}$ linearizes the subgraph context, splits into
subsequences and feeds it to the model chunk-by-chunk.  Our model
alleviates this problem by efficiently encoding the KG information
with a graph neural network, preserving dependencies captured in its
structure. As a result, DCG$_{\mathcal{GL}}$ performs better on most
question types, including simple questions, and reasoning-style
questions. We further compare DCG$_{\mathcal{GL}}$ against a variant
which uses context-dependent type linking instead
(DCG$_{\mathcal{CL}}$). We find that context-dependent type linking is
slightly worse than global lookup which is expected given that it does
not have access to the full list of KG types.

 In general, we observe that results with exact match (EM) are lower
 than F1 or Accuracy. EM is a stricter metric, it does not allow for
 any deviation from the goldstandard \textsc{Sparql}. However, it is
 possible for two queries to have different syntax but equivalent
 meaning, and for partially well-formed queries to evaluate to
 partially accurate results. In contrast to EM, F1 and Accuracy give
 partial credit and thus obtain higher scores.

\paragraph{The Effect of Entity Disambiguation}
We now present results with a variant of our model which operates over
\emph{disambiguated entities} (compare second and fifth blocks in
Table~\ref{tab:mainresult}, with heading DCG$_{\mathcal{CL}}$).  We
observe that disambiguation has a significant effect on model
performance, leading to an F1 increase of more than~11\%. We further
assess the utility of context-dependent linking by comparing
DCG$_{\mathcal{CL}}$ to a variant which does not have access to the
type graph~$G^{type}_{t}$, neither during training nor during
evaluation (see column DCG, With Disambiguation). This type-deficient
model performs overall worse both in terms of F1 and EM, but is still
superior to BertSP$_{\mathcal{GL}}$, even though the latter has access
to more information via the top-$K$ entity lookup and global type
linking. This points to the importance of encoding context in a
targeted manner rather than brute force. In
Appendix~\ref{appendix:context_len_appendix} we discuss the effect of
context length on the availability of type information.

\begin{figure}[t]
  \centering
    \begin{tikzpicture}[scale=0.76]
    \begin{axis}[xlabel={Turn Position},ylabel={Exact Match
          Accuracy (\%)},tick pos=left, legend
        style={at={(0.5,.88)},anchor=west},xtick={2,4,6,8,10,12,14,16,18},ytick
        distance=0.2, ymax=1] 
    \addplot table[x index=0,y index=4,col sep=comma] {dataplot.dat};
    \addlegendentry{BertSP$_{\mathcal{GL}}$ ($-$D)}

      \addplot[red,mark=triangle*, mark size=3pt] table[x index=0,y index=3,col sep=comma] {dataplot.dat};
    \addlegendentry{DCG$_{\mathcal{GL}}$ ($-$D)} 

     \addplot[brown,mark=triangle*, mark size=3pt] table[x index=0,y index=2,col sep=comma] {dataplot.dat};
    \addlegendentry{DCG$_{\mathcal{CL}}$ ($-$D)}
    
    \addplot[black,mark=triangle*, mark size=3pt] table[x index=0,y index=1,col sep=comma] {dataplot.dat};
    \addlegendentry{DCG$_{\mathcal{CL}}$ ($+$D)}
    
    \end{axis}
    \end{tikzpicture}
        \caption{Exact Match Accuracy averaged across question types
          at different turn positions. \mbox{$+/-$D} denotes the
          presence/absence of entity disambiguation.
    \label{fig:position_acc}}%
\end{figure}

\paragraph{The Effect of Conversation Length}
We next examine the benefits of modeling context dynamically.

Ideally, a model should produce an accurate semantic parse no matter
the conversation length.  Figure~\ref{fig:position_acc} plots exact
match accuracy (averaged across question types) against different turn
positions.  In general, we observe that utterances occurring later in
the conversation are more difficult to parse. As the dialogue
progresses, subsequent turns become more challenging for the model
which is expected to leverage the common ground established so far. This
involves maintaining the subgraph context based on the conversation
history in addition to handling linguistic phenomena such as
coreference and ellipsis. Overall, DCG$_{\mathcal{CL}}$ (with and
without entity disambiguation) is superior to BertSP$_{\mathcal{GL}}$,
and the gap between the two models is more pronounced for later turns.

\begin{table}[t]
  \centering
  \small
    \begin{tabular}{@{}l@{}|c@{~}c@{~}@{~}c@{~}|@{~}c@{}c@{}} \toprule
\multicolumn{1}{c}{}     & \multicolumn{3}{c}{Without Disambiguation} & \multicolumn{2}{@{}c@{}}{\hspace*{-.2cm}With Disambiguation}  \\
 \multicolumn{1}{c|}{}    & DCG$_{\mathcal{CL}}$ & DCG$_{\mathcal{GL}}$ &  BertSP$_{\mathcal{GL}}$ &  DCG$_{\mathcal{CL}}$ & DCG  \\
            \hline
            Coref=$-1$          &  {58.25} & 51.10 & 49.39 & \textbf{74.23}  & 72.22    \\
             Coref$<-1$         &  {23.52} & 12.42 & 00.00 & \textbf{33.64}  & 32.40         \\
            Ellipsis            &  {39.30} & {39.90} & 26.39 & \textbf{62.26}  & 61.10   \\
            MEntities           &  43.71 & {53.46} & 41.64 & \textbf{61.59}  & 58.90  \\
            \bottomrule
    \end{tabular}
      \caption{Average Exact Match accuracy.  Coref=$-1$ are
        utterances with referring expressions resolved in the
        previous turn. Coref$<-1$ are utterances with referring
        expressions  resolved in the wider discourse
        context, beyond the previous turn. MEntities abbreviates
        multiple entities and refers to utterances with plural
        mentions.} \label{tab:phenomena}
\end{table}

\paragraph{Modeling Discourse Phenomena}
\label{sec:linguistic}
Discourse phenomena, such as ellipsis and coreference, are prevalent
in conversations.  Ellipsis refers to grammatical omissions from an
utterance that can be recovered from context. In the
interaction:
\begin{examples}
\begin{small}
\item[Q1] What does Andrei Neagoe do for a living? 
\item[Q2] And  how about Wilhelm Dietrichson?
\end{small}
\end{examples}
the phrase \textsl{do for a living}
is elided from~(Q2) but can be understood in the context
of~(Q1). Coreference on the other hand, occurs between utterances that
refer to the same entity. For example, between utterances~2 and~3 in
Figure~\ref{fig:example_interaction}.


In Table~\ref{tab:phenomena}, using exact match, we assess how
different models handle ellipsis and coreference across question
types. Coref=$-1$ refers to cases where coreference can be resolved in
the immediate context, i.e.,~the previous turn. Coref$<-1$ involves
utterances that require access to wider conversation context, beyond
the previous turn. In the setting that does not disambiguate entities,
we observe that models which exploit discourse context (variants
DCG$_{\mathcal{GL}}$ and DCG$_{\mathcal{CL}}$) are better at resolving
co-referring and elliptical expressions compared to
BertSP$_{\mathcal{GL}}$. We also see that entity disambiguation is
very helpful, leading to substantial improvements for
DCG$_{\mathcal{CL}}$ across  discourse-related phenomena.

Similar to~\citet{spice}, we also evaluate model performance on
utterances with plural mentions; these are typically linked to
multiple entities which the semantic parser must enumerate in order to
build a correct parse (MEntities in Table~\ref{tab:phenomena}).
DCG$_{\mathcal{CL}}$ with disambiguation is overall best, while DCG
(without type linking) is worse. This is not surprising, utterances
with multiple entities generally have complex parses, with multiple
sub-queries and entity types, which DCG does not have access to.

 \paragraph{The Nature of Parsing Errors} Overall, we find that our
 model is able to predict syntactically valid \textsc{Sparql}
 queries. Errors are mostly due to misinterpretations of the
 question's intent given the graph context and previous questions or
 missing information. Our model also has difficulty parsing
 Clarification and Quantitative Reasoning questions.  For
 Clarification questions, it is not able to select the right entity
 after clarification. For example, in the following conversation:
 
\begin{table}[H]
 \begin{small}
 \begin{tabular}{l@{~}p{5.5cm}}
Answer: &  Peter G. Van Winkle, Arthur I. Boreman,
  William E. Stevenson\\
Utterance: & Which language
    is that person capable of writing ?\\
Clarification: &  Did you mean Arthur I. Boreman~? \\
Answer: & No, I meant Peter G. Van Winkle. Could
you tell me the answer for that?\\
 \end{tabular}
 \end{small}
\end{table}
\hspace*{-2.2ex}it selects \emph{Arthur I. Boreman} ({\tt Q709961})
instead of \emph{Peter G. Van Winkle} ({\tt Q1404201}) leading to an
incorrect SQL parse. In this case, the broader context overrides
useful information in the immediately preceding turn.  Determining
relevant context based on specific question intents would be helpful,
however, we leave this to future work.

Failures in type linking are a major cause of errors for Quantitative
Reasoning questions which typically have no or very limited context
(e.g.,~``Which railway stations were managed by exactly 1 social group
?''). However, our model relies on the availability of types in the
entity neighborhood, as it performs type linking in a context
dependent manner. We observe that it becomes better at parsing such
questions when given access to all KG types (see
Table~\ref{tab:mainresult}, DCG$_{\mathcal{GL}}$
vs. DCG$_{\mathcal{CL}}$).

\section{Conclusions}

In this paper, we present a semantic parser for KBQA which
interactively maps user utterances into executable logical forms, in
the context of previous utterances. Our model represents information
about utterances and their context as KG subgraphs which are created
dynamically and encoded using a graph neural network. We further
propose a context-dependent approach to type linking which is
efficient and scalable.

Our experiments reveal that better modeling of contextual information
improves performance, in terms of entity and type linking, resolving
coreference and ellipsis, and keeping track of the interaction history
as the conversation evolves. Directions for future work are many and
varied.  In experiments, we use an off-the shelf NER system, however,
jointly learning a semantic parser and an entity linker would be
mutually beneficial, avoiding error propagation. Given that it is
prohibitive to encode the entire KG, we encode relevant subgraphs on
the fly. We could further explicitly model the relationship
between KG entities and question tokens which previous work has shown
is promising \cite{wang-etal-2020-rat}. Finally, it would be
interesting to adapt our model so as to handle non-i.i.d
generalization settings.

\section{Limitations}
Our model relies on a pre-trained NER module for entity linking. As
this module is trained and evaluated on specific datasets, its
performance may not generalize on unseen domains within
Wikidata. Moreover, we did not explicitly consider relations. We
assume that the correct information will be available which may not
always be the case. We focus on encoding KG structural information and
pass the learning of the interactions between the KG and the
linguistic utterances to the decoder. As shown in previous
work~\cite{zhang2022greaselm}, effectively combining KG information
with a language model can be mutually beneficial in the context of
question answering. However, it requires an extensive study in itself
to determine the task specific parametrization~\cite{wang2022gnn}.

\section*{Acknowledgements}
This work is supported in part by Huawei and the UKRI Centre for
Doctoral Training in Natural Language Processing (grant
EP/S022481/1). Lapata gratefully acknowledges the support the UK
Engineering and Physical Sciences Research Council (grant
EP/W002876/1).

\bibliography{anthology,custom}
\bibliographystyle{acl_natbib}


\appendix
\section{The Effect of Context Length}
\label{appendix:context_len_appendix}

Figure~\ref{fig:context_len_acc} plots the performance of
DCG$_{\mathcal{CL}}$ (disambiguation setting) against progressively
increasing context length $\in [1,10]$. We observe that access to
wider context is beneficial up to a point. Performance deteriorates
with very long contexts (beyond turn position~5). We stipulate two
reasons for this. Firstly, longer interactions might be long because
users ask about more than one entity or topic, in which case local
context might be sufficient to provide an answer. And secondly, longer
interactions might be genuinely confusing and noisy for annotators to
create, let alone models.

\begin{figure}[t]
        \centering
    \begin{tikzpicture}[scale=0.75]
    \begin{axis}[xlabel={Turn Position},ylabel={Exact Match
          Accuracy (\%)},,tick pos=left, legend
        style={at={(0.5,.15)},anchor=west},xtick={1,3,5,7,9,10}]
    \addplot[black,mark=triangle*, mark size=3pt] table[x index=0,y index=1,col sep=comma] {datalastn.dat};
    \addlegendentry{DCG$_{\mathcal{CL}}$ ($+$D)}
    
    \end{axis}
    \end{tikzpicture}
        \caption{Exact Match Accuracy for different context lengths
          averaged across question types (validation set; $+$D:~with
          entity disambiguation).
    \label{fig:context_len_acc}}%
\end{figure}

\begin{table*}[t]
\small{
    \resizebox{\textwidth}{!}{\begin{tabular}{@{}l|c@{~~}c|c@{~~}c|c@{~~}c|c@{~~}c|r@{~~}r|r@{~~}r@{}}
    \toprule
\multicolumn{1}{c}{}    & \multicolumn{4}{c}{Context Length~1} & \multicolumn{4}{c}{Context Length~5} &  \multicolumn{4}{c}{\multirow{2}{*}{Diff due to type linking}} \\
\multicolumn{1}{c|}{}    & \multicolumn{2}{c|}{DCG$_{\mathcal{CL}}$} & \multicolumn{2}{c|}{DCG} & \multicolumn{2}{c|}{DCG$_{\mathcal{CL}}$} & \multicolumn{2}{c|}{DCG} \\
    \hline
        Question Type & F1 & EM & F1 & EM & F1 & EM & F1 & EM & $\Delta_{\rm F1_1}$ & $\Delta_{\rm EM_1}$ & $\Delta_{\rm F1_5}$ & $\Delta_{\rm EM_5}$\\ 
        \hline
        Clarification & 75.66 & 68.61 & 52.22 & 53.87 & 82.01 & 74.82 & 82.03 & 72.1 & 23.44 & 14.74 & 0.02 & 2.72 \\ 
        Logical Reasoning & 92.59 & 77.07 & 86.9 & 67.94 & 93.95 & 79.52 & 93.33 & 78.19 & 5.69 & 9.13 & 0.62 & 1.33 \\ 
        Quantitative Reasoning & 36.1 & 13.74 & 30.91 & 11.04 & 59.83 & 31.17 & 56.66 & 28.66 & 5.19 & 2.7 & 3.17 & 2.51 \\ 
        Comparative Reasoning & 76.72 & 39.75 & 74.77 & 37.79 & 90.91 & 62.46 & 90.09 & 61.11 & 1.95 & 1.96 & 0.82 & 1.35 \\ 
        Simple Question (Coref) & 88.18 & 79.83 & 83.04 & 76.27 & 88.49 & 79.9 & 87.41 & 79.18 & 5.14 & 3.56 & 1.08 & 0.72 \\ 
        Simple Question (Direct) & 87.56 & 61.59 & 79.74 & 58.21 & 88.27 & 62.25 & 85.6 & 61.44 & 7.82 & 3.38 & 2.67 & 0.81 \\ 
        Simple Question (Ellipsis) & 80.38 & 81.75 & 74.2 & 76.84 & 79.08 & 83.87 & 84.35 & 82.45 & 6.18 & 4.91 & 5.27 & 1.42 \\
        \hline
        ~ & AC & EM & AC & EM & AC & EM & AC & EM & $\Delta_{\rm AC_1}$ & $\Delta_{\rm EM_1}$ & $\Delta_{\rm AC_5}$ & $\Delta_{\rm EM_5}$ \\ 
        \hline
        Verification (Boolean) & 88.02 & 61.45 & 82.15 & 48.24 & 87.41 & 66.32 & 86.75 & 63.66 & 5.87 & 13.21 & 0.66 & 2.66 \\ 
        Quantitative Reasoning (Count) & 69.41 & 65.34 & 65.16 & 60.58 & 75.2 & 70.84 & 72.96 & 69.02 & 4.25 & 4.76 & 2.24 & 1.82 \\ 
        Comparative Reasoning (Count) & 42.81 & 30.5 & 39.74 & 28.69 & 67.7 & 57.34 & 66.76 & 56.6 & 3.07 & 1.81 & 0.94 & 0.74 \\ \hline 
        Overall & 73.74 & 57.96 & 66.88 & 51.95 & 81.28 & 66.85 & 80.59 & 65.24 & 6.86 & 6.01 & 0.69 & 1.61 \\ \bottomrule
    \end{tabular}}
    \caption{Interaction of context length and type linking. $\Delta_{\rm F1_1}$ is the absolute difference in F1
      score between DCG$_{\mathcal{CL}}$ and DCG for context
      length~1. $\Delta_{\rm F1_5}$ is the absolute F1 difference for
      context length~5. \label{appendix:tab:type_diff} } }
\end{table*}

We further assess how context length interacts with the availability
of type information.  Table~\ref{appendix:tab:type_diff} shows the
difference in performance with and without explicit type linking at
context lengths~1 and~5. As described in
Section~\ref{sec:model_config}, DCG does not have explicit type
linking while DCG$_{\mathcal{CL}}$ uses context-dependent linking,
while both models apply entity disambiguation. $\Delta_{\rm F1_1}$ is
the absolute difference in F1 score between DCG$_{\mathcal{CL}}$ and
DCG for context length~$1$. Similarly, $\Delta_{\rm F1_5}$ denotes the
difference for context length $5$.

Overall, we find a significant drop in performance for context
length~1 compared to context length~5. This indicates that more type
information becomes available with increased context length. However,
performance varies with question types. Specifically, the exact match
difference is lot bigger for Clarification questions compared to
Quantitative Reasoning questions  which seem to require access to
larger KB subgraphs. 

\section{Model Details}
\label{sec:app:model_details}
Our model is implemented using PyTorch~\citep{NEURIPS2019_bdbca288}
and trained with the AdamW~\citep{loshchilov2018decoupled}
optimizer. It was trained with an A100 GPU with a batch size of~64 and
an initial learning rate of 0.001. AdamW~coefficients $\beta_1$ and
$\beta_2$ (used for computing running averages of gradient and its
square) were set to $0.9$ and $0.999$, respectively. W The weight
  decay coefficient was set to 0.01 for all experiments. 
  Hyperparameters were set based on initial experiments using a
  manually selected grid. We did not tune learning rate parameters. We
  choose the number of GATv2 and decoder layers from $[1,4]$ and found $2$
  to work best. Our SPARQL query server was deployed using
  Blazegraph.\footnote{\url{https://blazegraph.com/}} which uses only
  CPU-based resources and has access to 100G~of RAM.

  We use two attention heads with GATv2.  Specifically, let~$K$ denote
  the attention head as computed~\citep{gat2018graph} in
  Equation~(\ref{eqnatt}). The output of each head is
  concatenated as follows:

\begin{equation*}\label{eqnatt_k}
        h^{l} = \concat_{k=1}^K \sigma\left(\sum_{j\in\mathcal{N}_i}\alpha_{ij}^k{W}^{k}h_j^{l - 1}\right)
\end{equation*}
where $\parallel$ represents concatenation. $\alpha_{ij}^k$ are normalized attention coefficients computed by the $k$-th attention mechanism as in Equation~(\ref{attncoeff}).

Our graph is represented as an adjacency matrix. To create a
mini-batch, adjacency matrices are diagonally
stacked~\citep{Fey/Lenssen/2019}. This creates a combined graph that
holds multiple isolated subgraphs as shown below:
\begin{align*}
    \begin{split}\mathbf{A} = \begin{bmatrix} \mathbf{A}_1 & & \\ & \ddots & \\ & & \mathbf{A}_n \end{bmatrix} \end{split}
\end{align*}
where $n$~is the batch-size number of graphs. Node input ${H}$ and
target $\bar{H}$ features are simply concatenated in the node
dimension as follows:
\begin{equation*}
    \begin{split}
        \qquad \mathbf{H} = \begin{bmatrix} \mathbf{H}_1 \\ \vdots \\ \mathbf{H}_n \end{bmatrix}, \qquad \mathbf{\bar{H}} = \begin{bmatrix} \mathbf{\bar{H}}_1 \\ \vdots \\ \mathbf{\bar{H}}_n \end{bmatrix}.
    \end{split}
\end{equation*}
\onecolumn

\section{The $\mathbb{SPICE}$ Dataset: Question Types}
\label{sec:app:ques_types}
\begin{center}
\begin{table*}[h!]
  \resizebox{\textwidth}{!}{\begin{tabular}{  l l } 
    \toprule
    \multirow{4}{*}{Simple Question (Ellipsis)} & \textcolor{darkgray!70}{Utterance: Who created the design for Samus Aran?} \\
                    & \textcolor{darkgray!70}{Answer: Hiroji Kiyotake} \\
                    & Utterance: And how about The Dreamland Chronicles: Freedom Ridge?   \\ 
                    & Answer: Julian Gollop \\
    \hline
    \arrayrulecolor{gray!50}
    \multirow{2}{*}{Simple Question (Direct)}  & Utterance: Who starred in Mathias Kneissl ? \\ 
        & Answer: Rainer Werner Fassbinder, Volker Schlöndorff, Hanna Schygulla \\
    \hline
    \multirow{2}{*}{Simple Question (Coreferenced)} & Utterance: Who was the director of that work of art ?  \\
        & Answer: Reinhard Hauff \\
    \hline
    \multirow{2}{*}{Verification (Boolean)} & Utterance: Does Dubashi have that person as actor ?    \\
        & Answer: No \\    \hline
    \multirow{3}{*}{Logical Reasoning (All)} & Utterance: Which works of art are Rainer Werner Fassbinder or 
                 \\& Laura Esquivel a screenwriter of ?  \\
                 & Answer: The American Soldier, Lili Marleen, Love Is Colder Than Death... \\
    \hline
    \multirow{5}{*}{Clarification}  & Utterance: How many administrative territories or political territories
                         \\& did that  work of art originate ? 
                        \\ & Did you mean Lili Marleen ? 
                        \\ & No, I meant Querelle. Could you tell me the answer for that?    
                        \\ & Answer: 2 \\
    \arrayrulecolor{black}
    \hline
    \multirow{3}{*}{Quantitative Reasoning (Count)} &  Utterance: How many still waters are situated nearby Norway
        \\ & or Austria-Hungary ?  
        \\ & Answer: 2 \\
    \hline
    \multirow{3}{*}{Comparative Reasoning (Count)}  & Utterance: How many watercourses are more number of landscapes \\&  located on than Kafue River ? 
    \\ & Answer: 2 \\
    \hline
    \multirow{3}{*}{Comparative Reasoning (All)} & Utterance: Which political territories are located nearby lesser number  \\ &  of watercourses or bodies of water than Bareyo ?   
    \\ & Answer: 2 \\
    \bottomrule
  \end{tabular}}
  \caption{Examples of question types attested in  $\mathbb{SPICE}$
    \cite{spice}. The utterance in gray is provided for ease of interpretation.} 
  \end{table*}
  \end{center}

\end{document}